# Object Detection using Oriented Window Learning Vision Transformer: Roadway Assets Recognition

Taqwa Alhadidi[a'], Ahmed Jaber[b], Shadi Jaradat[c], Huthifa Ashqar[d], Mohammed Elhenawy[c]

[a] Civil Engineering Department, Al-Ahliyya Amman University, Amman, Jordan, 19328.
t.alhadidi@ammanu.edu.jo

[b] Department of Transport Technology and Economics, Faculty of Transportation Engineering and Vehicle Engineering, Budapest University of Technology and Economics, Műegyetem rkp. 3., H-1111 Budapest, Hungary

[c] Accident Research and Road Safety Queensland, Queensland University of Technology, Brisbane, 130 Victoria Park Rd, Kelvin Grove QLD 4059, Australia

[d] Arab American University, Jenin, Palestine and Columbia University, NY, USA

**Abstract.** Object detection is a critical component of transportation systems, particularly for applications such as autonomous driving, traffic monitoring, and infrastructure maintenance. Traditional object detection methods often struggle with limited data and variability in object appearance. The Oriented Window Learning Vision Transformer (OWL-ViT) offers a novel approach by adapting window orientations to the geometry and existence of objects, making it highly suitable for detecting diverse roadway assets. This study leverages OWL-ViT within a one-shot learning framework to recognize transportation infrastructure components, such as traffic signs, poles, pavement, and cracks. This study presents a novel method for roadway asset detection using OWL-ViT. We conducted a series of experiments to evaluate the performance of the model in terms of detection consistency, semantic flexibility, visual context adaptability, resolution robustness, and impact of non-max suppression. The results demonstrate the high efficiency and reliability of the OWL-ViT across various scenarios, underscoring its potential to enhance the safety and efficiency of intelligent transportation systems.

## 1 Introduction

Object detection plays a vital role in transportation applications, such as surveillance, traffic monitoring, and autonomous driving. Implemented in several studies, object detection shows its importance in Advanced Driving Assistant systems (ADAS)[1], and enhances the reliability and safety of autonomous driving systems[2]. Conventional object detection methods usually rely on large-labeled datasets and face challenges when dealing with limited data situations that are common in specialized infrastructure aspects. Technologies such as LiDAR-based SLAM systems [3] and LiDAR sensors [4] can act as enablers for intelligent transportation systems and vehicles. As such, these advanced perception technologies can realize the real-time detection and location of objects, providing rich real-time perception information for decision-making in various traffic scenes. Vision transformers present a possible method to improve these systems by offering efficient object detecting capabilities. The incorporation of transformers in transportation corresponds to the wider pattern of employing transformers in diverse computer vision applications, such as picture classification, action identification, and segmentation [5, 6].

Owing to the advantages of high detection efficiency and accuracy, one-shot object detection technology can solve the problem effectively in the field of transportation [7]. It easily identifies and localizes vehicles, pedestrians, and obstacles in various scenarios[1] . One-shot object detection technologies are popular in transportation, using one labelled query image to test these approaches with little labelled data. Standard object detection architectures are combined with Siamese backbones, feature attention methods, and concatenation techniques to improve one-shot object detection accuracy [8]. Researchers have tried many methods for one-shot object detection. Deep learning has been applied to improve low-shot object detection by finding items from query photographs based on support images of the same category[9] . Innovative methods like Laplacian objects and rapid matrix cosine similarity improve generic one-shot object detection[10]. Furthermore, the concept of zero-shot object detection has been introduced to detect both seen and unseen classes during testing, providing a comprehensive approach to object detection in varying scenarios [11, 12]. Additionally, the idea of any-shot object detection has been proposed to handle the detection of both zero-shot and few-shot object classes simultaneously[13].

The Oriented Window Learning Vision Transformer (OWL-ViT) presents a compelling approach by integrating adaptable windows that modify their orientation according to the object's existence and geometry[14]. Precisely capturing the various orientations and sizes of items such as cracks, poles, and manholes from a small number of instances is of utmost importance. It is employed as an image encoder to produce features and align them with a large language model[15].

This study aims to utilise OWL-ViT to improve the detection of transport infrastructure inside a one-shot learning framework. Our objective is to demonstrate the excellent performance of OWL-ViT in transportation applications by customising it to meet their specific requirements. This paper provides a comprehensive account of our methodology, the meticulous planning of our experiments, and the notable progress achieved in implementing OWL-ViT for practical transportation monitoring and maintenance in real-world scenarios. The key contribution of this paper is as follows:

1- Test the ability to implement OWL-ViT in transportation applications.
2- Test OWL-ViT performance in detecting different roadway assets in zero-shot.
3- Test the effect of image lightening and resolution on the OWL-ViT performance.
4- The effect of non max suppression of the performance of OWL-ViT.

To the best of our knowledge, this paper is the first to address not only the application of OWL-ViT in transportation asset detection but also it covers reporting the performance of OWL-ViT in various scenarios to detect multiple objects simultaneously [15–18]. The limitation of our work is that there is currently no extensive transportation data available that has the ground truth to recognize multiple roadway assets to compare the OWL-ViT performance, however, we evaluated the model's performance by comparing the manual inspection results to the OWL-ViT results.

## 2 Methodology

In this section, we present the adopted research methodology. A description with the description of the Oriented Window Learning Vision Transformer (OWL-VIT), describing the research experiment questions, and the proposed research methodology.

### 2.1 OWL-ViT Architecture

We provide a concise overview of the OWL-ViT model, which serves as our detection backbone. OWL-ViT comprises a conventional Vision Transformer[16] for encoding images and a text encoder that follows a similar architectural design. The encoders undergo contrastive pretraining using extensive datasets consisting of pairs of images and corresponding texts[17] . Following the pretraining phase, the model is transitioned to detection by incorporating additional classification and box regression heads. These heads are responsible for predicting class embeddings and box coordinates directly from the image encoder output tokens. In open-vocabulary classification, similarities are calculated by taking the inner product between class embeddings obtained from image patches and text embeddings of label names, which are given as text prompts. The classification logits in question are modified by shifting and scaling them with learned parameters. They are then trained using a sigmoid focused loss on commonly used detection datasets. Model architecture is shown in Figure 1.

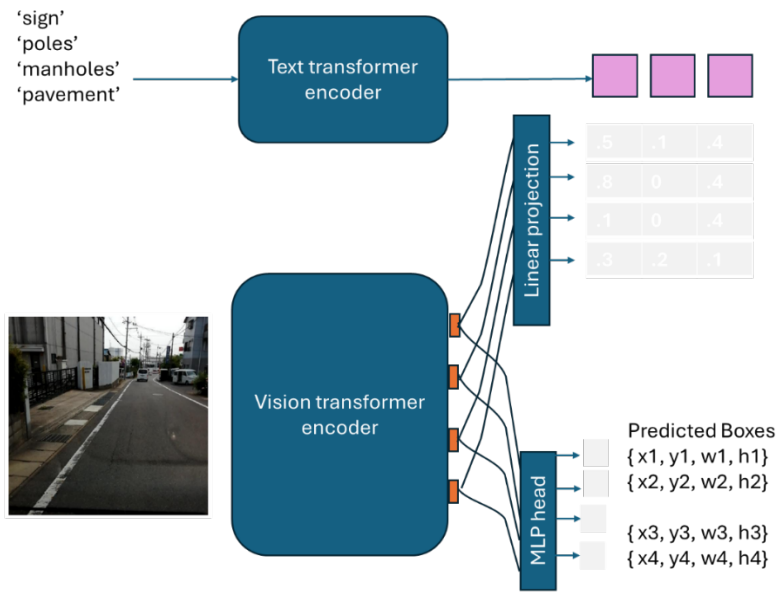

**Fig 1.** OWL-ViT Architecture

### 2.2 Encoder-Decoder OWL-ViT

To ensure that object representations can accurately follow objects across frames, it is necessary to separate object representations from individual image tokens. To achieve this, we incorporate a Transformer decoder between the encoder and the object heads, specifically the readout heads for box and class prediction, using the same approach as the original DETR architecture[18]. The decoder queries act as "slots" that recurrently carry object representations from one time step to the next. This architecture is commonly referred to as Enc-dec OWL-ViT. Given that OWL-ViT was originally developed as an architecture that merely encodes information, a crucial question arises regarding the preservation of the model's ability to function effectively with the addition of a decoder while accommodating an open vocabulary [19].

## 2.3 Experimental Design

Our research objective is to not only to implement the OWL-ViT in transportation asset recognition only but to check the performance of OWL-ViT across different scenarios. To do so, we have designed a series of experiments, each tailored to address specific research questions, these experiments are discussed below:

- **Experiment 1: Results Over Repeated Trials**

Our primary objective in this preliminary investigation is to assess the stability of OWL-ViT results by performing ten separate trials using a single image with a fixed text prompt. This process will enable us to gauge the model's capacity for maintaining consistent detection accuracy in identical input circumstances, thereby offering invaluable insights into its dependability for practical applications in various fields.

- **Experiment 2: Semantic Flexibility**

In order to evaluate the semantic adaptability of OWL-ViT, we conducted an experiment that employed a variety of synonyms within the query text for the same image. This study aimed to assess the model's ability to comprehend contextually related yet textually dissimilar queries while maintaining a high level of detection accuracy, thereby highlighting its potential to process a wide range of linguistically diverse inputs.

- **Experiment 3: Consistency Across Visual Contexts**

This study evaluates the model's uniformity in identifying query text elements in various visual contexts. By employing the identical textual query in distinct visual circumstances, we aim to determine whether OWL-ViT displays consistent recognition and precise detection of the specified components, regardless of the underlying image composition.

- **Experiment 4: Resolution Robustness**

We also evaluate the performance of the model across different image resolutions. This experiment tests the efficacy of OWL-ViT by subjecting it to the same text query on images of different resolutions and different angles, in order to determine if image resolution has an impact on the accuracy of object detection. This is a crucial factor for applications in environments with fluctuating image qualities.

- **Experiment 5: Impact of Non-Max Suppression**

The last experiment examines the impact of non-max suppression, a method that enhances the detection process by eradicating unnecessary bounding boxes. By implementing non-max suppression in various images, we aim to determine if the model's precision and effectiveness can be improved, resulting in more precise and accurate outcomes.

## 2.4 Research Questions

In summary, we aim to respond on the following research questions:

1. Does OWL-VIT maintain the same detection accuracy after we run the same prompt using multiple runs?

2. Does OWL-VIT maintain the same detection accuracy after we use different word synonyms in the query text on the same image?
3. Does OWL-VIT maintain the same detection accuracy for all query text elements?
4. Does OWL-VIT maintain the same detection accuracy for all query text elements under different image resolutions?
5. Does the result of OWL-ViT change if we use the non-max suppression?

To answer these questions, the adopted methodology is shown in Figure 2.

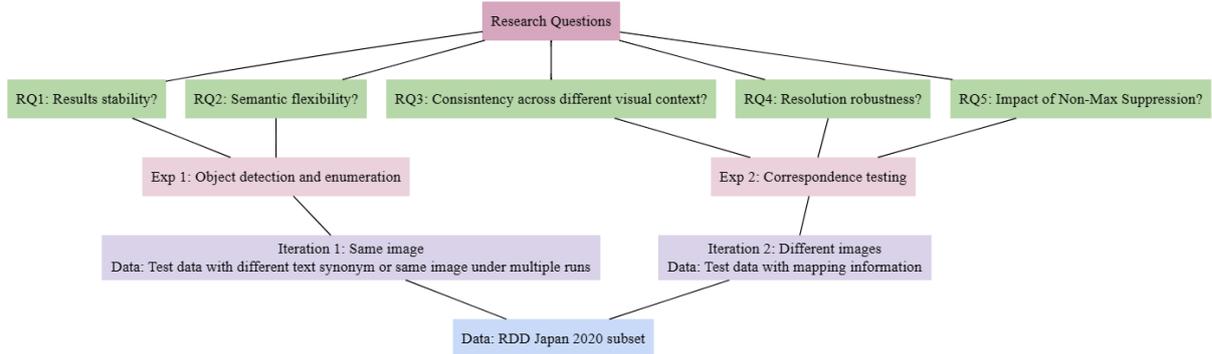

**Fig 2.** Proposed Research Methodology

### 2.5 Description of the Dataset

This study utilizes the recently updated 2022 dataset obtained from the Road Damage Detection Challenge. The dataset consists of road photographs gathered from different sites throughout Japan. This dataset is crucial for assessing the efficacy of the OWL-ViT (Oriented Window Learning Vision Transformer) model in accurately detecting and categorizing various forms of road damage across varying environmental circumstances. The dataset contains a vast assortment of road photos, with each image having precise bounding boxes that show the existence and exact location of road damage. The photos are categorized based on several types of damage, including cracks, potholes, and other surface flaws. These categories represent the typical road conditions found throughout Japan. The dataset's wide range of geographical and environmental characteristics provides a strong and reliable platform for testing our model. The ground truth of the data was not reported, as such we ran the model on a subset of the data.

## 3 Results and Discussions

### 3.1 Experiment 1: Results Over Repeated Runs

In this experiment we ran a prompt using the same picture multiple times in order to check the OWL-ViT results. The query text includes traffic signs, traffic pole, pothole, pavement, and pavement cracks. Results of the different trials for the same image are shown in Figure 3.

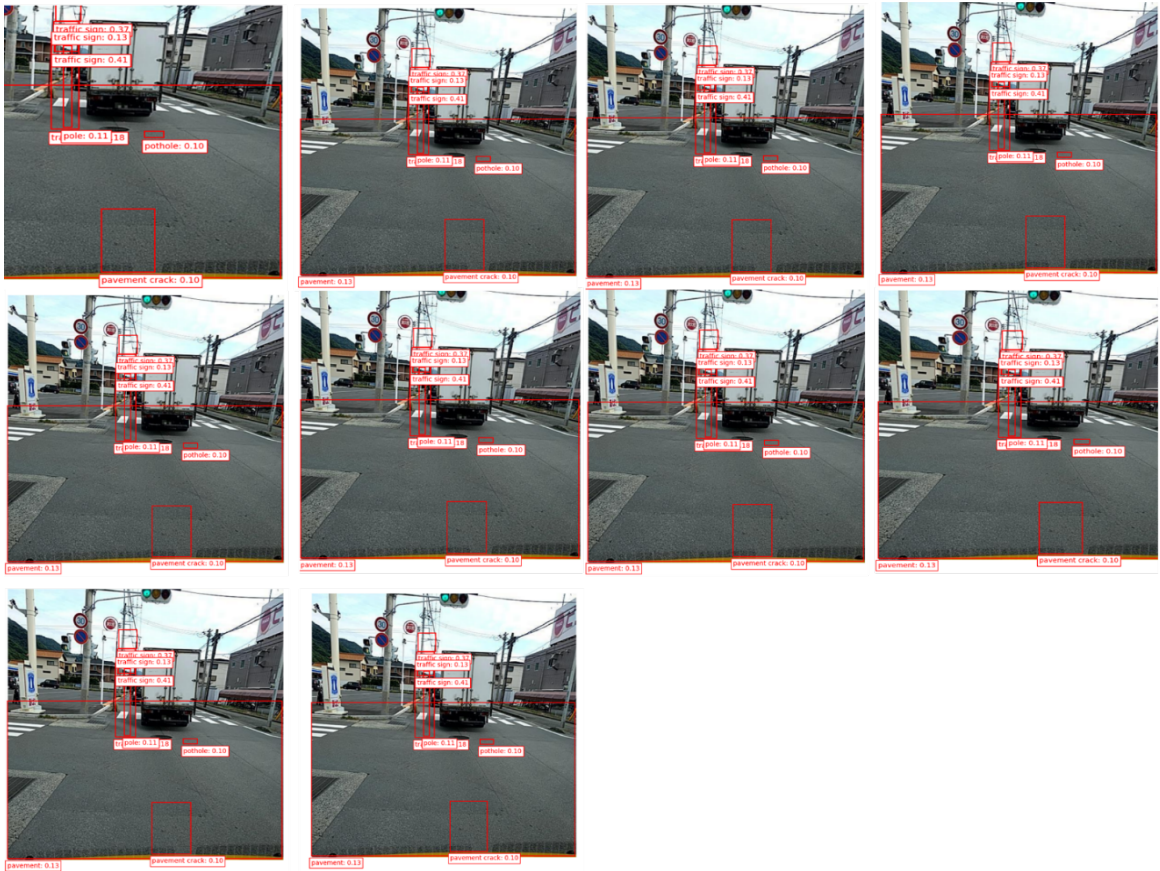

**Fig 3.** Results of the different trials for the same image

Analyzing Figure 3, the OWL-ViT model has successfully identified three distinct traffic signs within each image, with confidence scores demonstrating a notable level of certainty, particularly for one sign that consistently earns a higher score. This finding highlights the model's exceptional precision in recognizing traffic signs from various perspectives and distances. The results suggest that the model has identified the traffic sign pole with a lower confidence score, indicating that while it recognizes the pole, it may experience some uncertainty regarding its classification, potentially due to its resemblance to other vertical elements present in urban scenes. The model has maintained a moderate confidence score in detecting potholes, which might be attributed to their smaller size, or less distinctive features compared to other objects. However, for pavement and pavement cracks, despite the model's consistent detection of these elements across different trials, it has earned lower confidence scores, potentially reflecting challenges in differentiating them from other similar textures or surfaces in urban settings. Summarizing the different detected objects with their confidence scores across the multiple runs are shown in Table 1.

Table 1. Different run summary.

| Object | Run 1 | Run 2 | Run 3 | Run 4 | Run 5 | Run 6 | Run 7 | Run 8 | Run 9 | Run 10 | Average Confidence |
|---|---|---|---|---|---|---|---|---|---|---|---|
| **Traffic Sign 1** | 0.37 | 0.37 | 0.37 | 0.37 | 0.37 | 0.37 | 0.37 | 0.37 | 0.37 | 0.37 | 0.37 |
| **Traffic Sign 2** | 0.13 | 0.13 | 0.13 | 0.13 | 0.13 | 0.13 | 0.13 | 0.13 | 0.13 | 0.13 | 0.13 |
| **Traffic Sign 3** | 0.41 | 0.41 | 0.41 | 0.41 | 0.41 | 0.41 | 0.41 | 0.41 | 0.41 | 0.41 | 0.41 |
| **Traffic Pole** | 0.11 | 0.11 | 0.11 | 0.11 | 0.11 | 0.11 | 0.11 | 0.11 | 0.11 | 0.11 | 0.11 |
| **Pothole** | 0.1 | 0.1 | 0.1 | 0.1 | 0.1 | 0.1 | 0.1 | 0.1 | 0.1 | 0.1 | 0.1 |
| **Pavement** | 0.13 | 0.13 | 0.13 | 0.13 | 0.13 | 0.13 | 0.13 | 0.13 | 0.13 | 0.13 | 0.13 |
| **Pavement Crack** | 0.1 | 0.1 | 0.1 | 0.1 | 0.1 | 0.1 | 0.1 | 0.1 | 0.1 | 0.1 | 0.1 |

### 3.2 Experiment 2: Semantic Flexibility

To address Research Question 2, which examines whether the OWL-ViT model maintains consistent detection accuracy when exposed to variations in synonym usage within text queries, we conducted an analysis comparing outcomes across two different sets of queries aimed at identifying the same objects in the same image. The initial set of queries included terms like "pavement distress," while the second set used slightly different terms such as "pavement crack.". Output of the two different prompts are shown in figure 4 below.

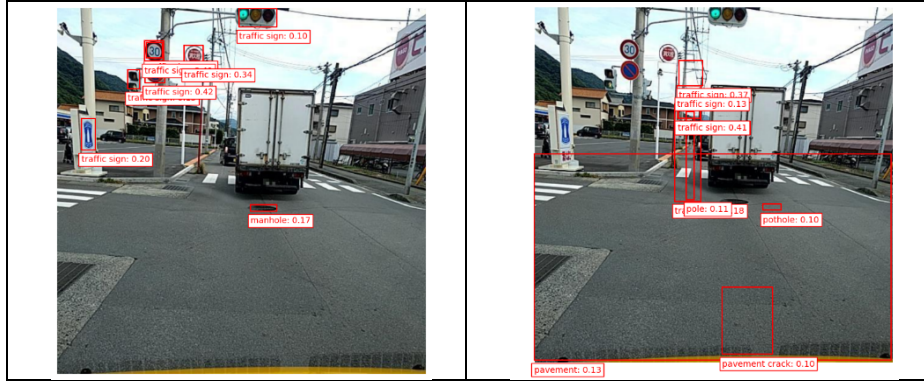

**Fig 4.** Results of Different Text Quaries. (a) : Text prompt text_queries = ["distressed pavement","pavement","manhole",'pole','traffic sign','pothole','sidewalk', (b): text_queries1 = ["pavement crack","pavement","manhole",'pole','traffic sign','pothole','crack','crosswalk']

Our study uncovered a high level of dependability in the model's ability to identify straightforward objects, such as traffic signs, poles, and potholes, as evidenced by the consistent results across both query sets in Figure 4. However, we observed a notable variation in the detection of "pavement" and "pavement crack," which were only identified in the second set of queries. This disparity indicates that the model's performance in recognizing more intricate or variably described features is impacted by the specific language used in the text queries. This finding points to an area for improvement in the model's natural language processing capabilities, as its sensitivity to terminology suggests room for enhancement. Summary of the model findings in both text query is presented in Table 2.

Table 2. Different Semantic Query Results

| Object | Query Term in Prompt 1 | Confidence in Prompt 1 | Query Term in Prompt 2 | Confidence in Prompt 2 |
|---|---|---|---|---|
| **Traffic Signs** | traffic sign | 0.37, 0.13, 0.41 | traffic sign | 0.37, 0.13, 0.41 |
| **Traffic Pole** | pole | 0.11 | pole | 0.11 |
| **Pothole** | pothole | 0.1 | pothole | 0.1 |
| **Pavement** | pavement | Not Detected | pavement | 0.13 |
| **Pavement Crack** | crack | Not Detected | pavement crack | 0.1 |

### 3.3 Experiments 3, 4, and 5

In order to answer the rest of the questions, we ran the following prompt on 20 random images from the RDD Japan Dataset. Data and code can be found using this link. The selected images cover different driving conditions, roadway characteristics, lightening, resolutions, and camera angles.

text_queries = ["distressed pavement", "pavement", "manhole", 'pole', 'traffic sign', 'pothole', 'pavement crack','pavement rutting','aligator pavement','sidewalk']

Sample of these images are shown in Figure 5.

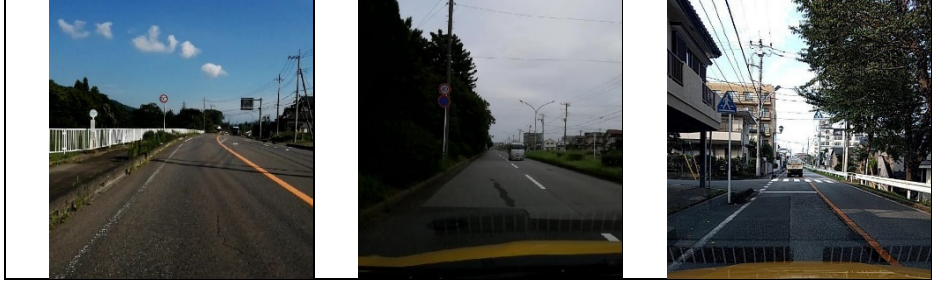

**Fig 5.** Sample of different images

In our study, we identified notable differences in the outcomes of the OWL-ViT model's object detection across a sample of 20 images. Specifically, we observed inconsistency in the model's ability to accurately detect specific elements, such as potholes and traffic signs. Additionally, we found that factors intrinsic to the images, including resolution, lighting conditions, and scene composition, appeared to impact the model's performance. These findings indicate a sensitivity in the OWL-ViT model to subtle variations in visual input, which may affect its ability to generalize across different real-world scenarios. Out of the 20 images, two images were not processed at all, few objects were detected in some images, multiple objects were detected correctly, or inaccurate bounded box. Sample of the different detection scores in different images are shown in Figure 6.

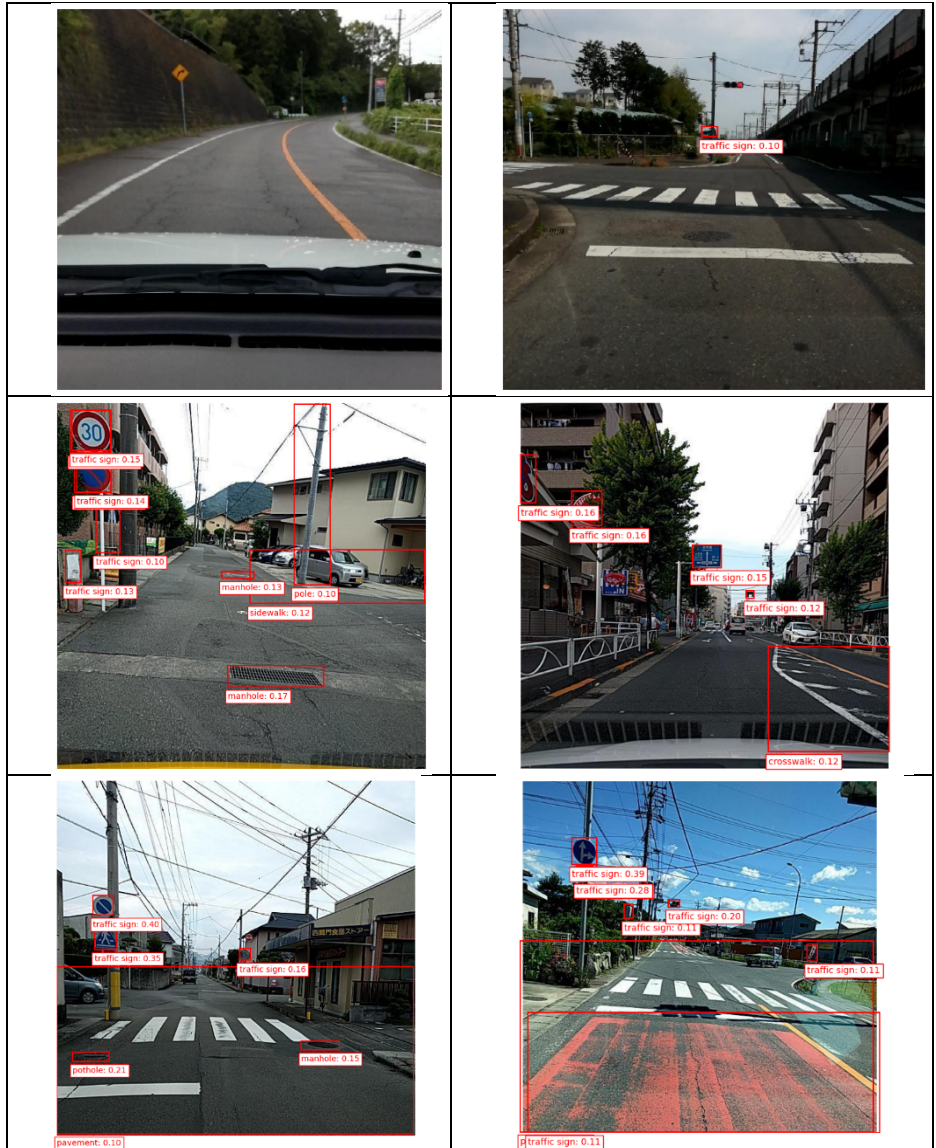

**Fig 6.** Sample of the different detection scores in different images

A total of 10,369 objects were detected from the 18 images with different confidence scores. The minimum confidence score was zero and the maximum confidence score was 0.5204. A summary of the different text query elements detection results with their statistics are presented in Table 3.

**Table 3:** Detection Results with no trimming

| Object Name | count | mean | std | min | 25% | 50% | 75% | max |
|---|---|---|---|---|---|---|---|---|
| alligator pavement | 134 | 0.0082 | 0.0093 | 0.0000 | 0.0006 | 0.0067 | 0.0117 | 0.0616 |
| distressed pavement | 108 | 0.0043 | 0.0058 | 0.0000 | 0.0004 | 0.0019 | 0.0057 | 0.0262 |
| manhole | 351 | 0.0169 | 0.0322 | 0.0001 | 0.0034 | 0.0080 | 0.0170 | 0.3428 |
| pavement | 919 | 0.0087 | 0.0145 | 0.0000 | 0.0005 | 0.0027 | 0.0122 | 0.1756 |
| pavement crack | 1,983 | 0.0066 | 0.0119 | 0.0000 | 0.0003 | 0.0017 | 0.0076 | 0.1112 |
| pavement rutting | 67 | 0.0003 | 0.0007 | 0.0000 | 0.0000 | 0.0001 | 0.0002 | 0.0035 |
| pole | 923 | 0.0086 | 0.0165 | 0.0000 | 0.0007 | 0.0025 | 0.0078 | 0.1302 |
| pothole | 504 | 0.0081 | 0.0157 | 0.0000 | 0.0010 | 0.0039 | 0.0087 | 0.2076 |
| sidewalk | 987 | 0.0096 | 0.0148 | 0.0000 | 0.0007 | 0.0036 | 0.0119 | 0.1262 |
| traffic sign | 4,392 | 0.0144 | 0.0356 | 0.0000 | 0.0015 | 0.0050 | 0.0136 | 0.5205 |

The analysis of the object detection scores from the dataset reveals significant variations in detection accuracy and reliability across different types of road-related objects. Traffic signs, with the highest count of detections (4,392), exhibit a broad range of detection scores, from 0.0000 to 0.5205, indicating that some detections are extremely reliable, although the overall average score remains low at 0.0144. This variability suggests that the detection system performs well in some conditions but may struggle in others. Pavement cracks, another commonly detected object, show similar trends, with a large number of detections (1,983) but lower average detection scores, which might indicate difficulty in consistently recognizing smaller or less obvious cracks. Sidewalks and general pavement surfaces also show moderate detection frequencies with average scores of 0.0096 and 0.0087, respectively, pointing to a decent capability in monitoring pedestrian pathways and pavement conditions, yet the variation in scores suggests that accuracy might not be consistent across different environments or conditions. Utility-related objects such as poles and manholes are detected fairly frequently, with manholes showing relatively higher detection confidence, essential for safety and maintenance planning. The detection of potholes and specific pavement distress types like distressed and alligator pavement, although less frequent, highlights critical areas needing maintenance. However, the lower detection scores for pavement rutting, despite its importance for vehicle stability, indicate potential challenges in identifying such damage accurately. Given the fact that the ground truth of this dataset is not reported, the model performance evaluation was done using visual inspection. In order to facilitate the process, the detection box was drawn is the detection score equals 0.1 or more. The reduced detection results are shown in Table 4.

**Table 4:** Detection Results with threshold trimming

| Object name | count | mean | std | min | 0.2500 | 0.5000 | 0.7500 | max |
|---|---|---|---|---|---|---|---|---|
| Manhole | 7 | 0.2048 | 0.0723 | 0.1333 | 0.1609 | 0.1727 | 0.2313 | 0.3428 |
| Pavement | 5 | 0.1262 | 0.0290 | 0.1010 | 0.1123 | 0.1161 | 0.1259 | 0.1756 |
| Pavement crack | 3 | 0.1079 | 0.0057 | 0.1013 | 0.1062 | 0.1110 | 0.1111 | 0.1112 |
| Pole | 3 | 0.1148 | 0.0139 | 0.1034 | 0.1071 | 0.1108 | 0.1205 | 0.1302 |
| Pothole | 3 | 0.1401 | 0.0586 | 0.1044 | 0.1063 | 0.1081 | 0.1579 | 0.2076 |
| Sidewalk | 3 | 0.1170 | 0.0103 | 0.1059 | 0.1124 | 0.1189 | 0.1226 | 0.1262 |
| Traffic sign | 86 | 0.2101 | 0.1218 | 0.1002 | 0.1198 | 0.1591 | 0.2728 | 0.5205 |

By setting a detection score threshold of 0.1, a more in-depth study uncovers precise observations about the model's capacity to accurately recognize objects related to roads with a high level of confidence output of this threshold was presented in Figure 4(b) . This technique emphasizes detections that are likely to be more precise, offering a more distinct representation of the model's performance in real-life situations. The average confidence for detecting manholes is remarkably high at 0.2048, with the highest score of 0.3428 over 7 detections. This indicates a strong capacity to properly identify these characteristics. Although there were only 5 pavement observations, they displayed moderate reliability, with scores ranging up to 0.1756. However, there was less variability in these ratings. The occurrence of pavement cracks, although seen just three times, consistently displays a high degree of scoring, suggesting that the detection of these features is dependable when found. The presence of poles and pavements has been identified three times each, with confidence levels of approximately 0.1148 and 0.117, respectively. This demonstrates a robust detecting capacity, especially for pavements, which are crucial for ensuring pedestrian safety. Potholes, which were also identified three times, exhibit a broader spectrum of scores ranging from 0.1044 to 0.2076, indicating considerable variability that implies inconsistent detection accuracy depending on the conditions or attributes of the pothole. The traffic signs, which have the highest number of detections that are highly confident at 86, exhibit the most significant range in scores, ranging from 0.1002 to a substantial 0.5205. The wide range of detection accuracy for traffic signs suggests that certain signs are detected with high precision, while others only slightly exceed the confidence level. This variation in accuracy may be impacted by factors such as sign visibility, angle, and climatic conditions.

### 3.4 Non-Max Suppression

The application of non-max suppression (NMS) in the recognition of traffic signs can be examined based on the detection scores and bounding boxes shown in Figure 7.

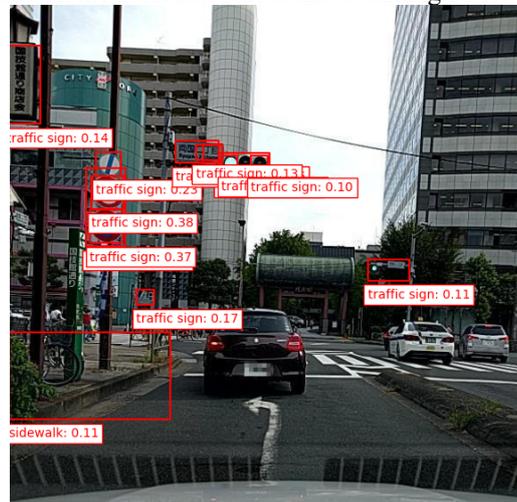

**Fig 7.** : Effect of the Non-max suppression on the OWL-ViT detection.

The takeaways from the image are:
- Multiple Detections for Certain Signs: There are several traffic signs detected in the image with varying scores. For example, one sign is detected with scores of 0.38, 0.37, and 0.25, indicating numerous potential boxes for the same object before suppression.
- Efficacy of NMS: Non-max suppression appears to have been implemented because there is no significant overlap of boxes. The boxes surrounding the traffic signs are singular and usually non-overlapping, demonstrating that NMS has effectively minimized duplication by preserving only the boxes with the highest confidence scores (e.g., scores of 0.38 and 0.37).
- Potential Constraints: Although effective, some slight overlaps or close proximities suggest that NMS might not entirely eliminate all sub-optimal boxes, especially when detection scores are close. This might occur when suppression thresholds are not appropriately established, or the model is imprecise.

In order to ease tracking the objects detection and reporting the OWL-ViT performance, manual inspection for each image was done as the ground truth is not there, summary of the manual inspection for the 20 images is shown in Table 5. The results show that there is a dispersion in the total numbers of the detected objects. Moreover, the OWL-ViT has not accurately detected sidewalk and pavement area Figure 7 shows some problem in OWL-ViT detection accuracy.

**Table 5:** Manual inspection results

| Image | Traffic Sign | Poles | Manhole | Pavement | Cracks | Alligator Crack | Sidewalk | Crossing | Pothole |
|---|---|---|---|---|---|---|---|---|---|
| 1 | 2 | 1 | 0 | 1 | 1 | 0 | 1 | 0 | 0 |
| 2 | 4 | 2 | 0 | 1 | 3 | 0 | 1 | 1 | 0 |
| 3 | 1 | 1 | 1 | 1 |   | 1 | 0 | 0 | 1 |
| 4 | 2 | 2 | 0 | 1 | 3 | 0 | 1 | 0 | 1 |
| 5 | 0 | 0 | 0 | 1 | 0 | 0 | 0 | 0 | 3 |
| 6 | 2 | 2 | 1 | 1 | 1 |   | 1 | 0 | 0 |
| 7 | 1 | 1 | 0 | 1 | 1 | 0 | 0 | 0 | 0 |
| 8 | 1 | 1 | 0 | 1 | 3 | 0 | 0 | 0 | 0 |
| 9 | 4 | 3 | 1 | 1 | 2 | 0 | 1 | 1 | 0 |
| 10 | 2 | 1 | 2 | 1 | 1 | 1 | 2 |   | 0 |
| 11 | 7 | 3 | 0 | 1 | 5 | 0 | 2 | 2 | 0 |
| 12 | 2 | 4 | 0 | 1 | 4 | 0 | 3 | 2 | 0 |
| 13 | 4 | 2 | 1 | 1 | 4 | 0 | 4 | 1 | 2 |
| 14 | 2 | 3 | 2 | 1 | 4 | 0 | 3 | 2 | 0 |
| 15 | 1 | 2 | 2 | 1 | 3 | 0 | 1 | 1 | 1 |
| 16 | 2 | 2 | 0 | 1 | 2 | 0 | 1 | 0 | 0 |
| 17 | 1 | 3 | 0 | 1 | 0 | 0 | 0 | 0 | 5 |
| 18 | 3 | 2 | 2 | 1 | 0 | 0 | 1 | 1 | 0 |
| 19 | 2 | 1 | 0 | 1 | 3 | 0 | 0 | 0 | 1 |
| 20 | 1 | 1 | 1 | 3 | 3 | 0 | 0 | 0 | 0 |

Figure 7 shows that there is a problem in several roadway assets detection results, this problem raises once there are variation in the object elevation as it can be seen from the figure. Moreover, if the crack occurs at the image corner, crack nor the pavement are detected. The evaluation metrics of the different roadway assets elements are presented in Table 6.

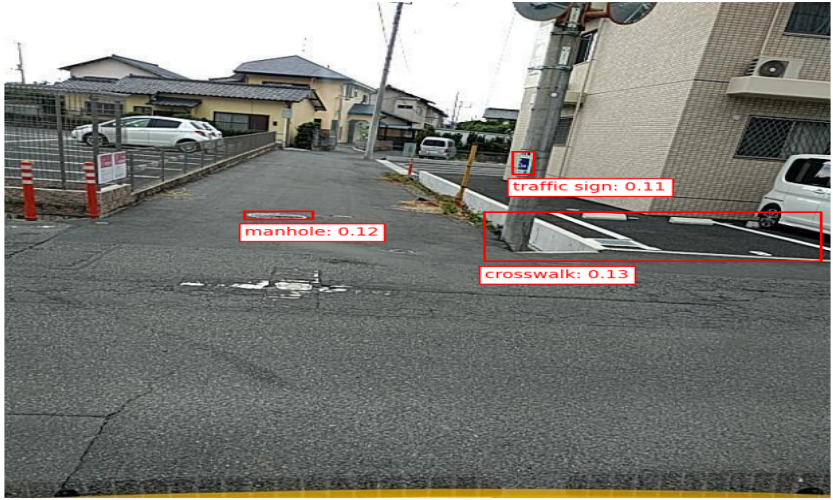

**Fig 7.** OWL-ViT detection accuracy results
**Table 5:** OWL-ViT Results

| Category | Precision | Recall | F1 Score |
|---|---|---|---|
| **Signs** | 83.33% | 94.59% | 88.61% |
| **Poles** | 83.33% | 41.67% | 55.56% |
| **Manhole** | 85.71% | 50.00% | 63.16% |
| **Pavement Cracks** | 60.00% | 30.00% | 40.00% |
| **Alligator Cracks** | 100.00% | 6.98% | 13.04% |
| **Sidewalk** | 50.00% | 4.76% | 8.70% |
| **Crossing** | 100.00% | 9.09% | 16.67% |

The results of the performance evaluation reveal strong the OWL-ViT detection capabilities for signs, with high recall and room for improvement in precision to reduce false positives. Precision is moderate for poles and manholes, while recall is lower, particularly for poles. Pavement cracks present challenges in accurate and reliable detection due to their varied appearances and conditions. Alligator cracks display perfect precision but extremely low recall, necessitating significant improvements in the detection algorithm. Sidewalk and crossing categories struggle with low recall and moderate to high precision, indicating a need for better tuning or more conservative detection to improve performance.

## 4  Discussion

The OWL-ViT model displays varying performance levels across different object categories in urban and roadway environments, highlighting both its strengths and areas for improvement in current deployment. For traffic signs, the model demonstrates exceptional effectiveness, achieving an impressive F1 score of 88.61%, thanks to high precision and recall. This indicates that the model is remarkably optimized for recognizing traffic signs, making it highly dependable for applications that require traffic sign detection, such as in automated driving systems or traffic management solutions.

However, the model exhibits limitations in accurately detecting and identifying other critical roadway elements. For instance, while the model demonstrates precise identification of poles, its recall is relatively low, suggesting that the model often misses many poles. This could impact applications where pole recognition is essential for navigation or obstacle avoidance.

The detection of manholes and pavement cracks shows a moderate level of accuracy but a lower detection rate. These results suggest a need for further training and model refinement, especially in capturing finer details or less prominent objects on roads, which are crucial for maintenance and safety assessments.

Of particular concern are the results for alligator cracks, sidewalks, and crossings, where the model displays very high precision but very low recall rates. This discrepancy implies that while the model can accurately identify these features when it detects them, it fails to recognize most instances. Such a characteristic could severely limit the model's usefulness in comprehensive road condition assessments and pedestrian safety measures, where overlooking these features could lead to safety hazards.

The detection of traffic signs in the image showcases the implementation of non-max suppression efficiently reducing the number of overlapping boxes and emphasizing the more confident detections. Adjustments to the NMS threshold could potentially lessen the remaining overlaps even further, enhancing the precision of the detection system. This analysis exhibits NMS's critical role in refining detection results, particularly in complex urban situations where several items can be adjacent to one another.

## 5    Conclusions

The OWL-ViT model has demonstrated considerable promise in detecting transportation infrastructure components, effectively addressing the challenges posed by limited data and variability in object appearances. Through our experiments, the model consistently performed well in identifying various traffic signs, highlighting its ability to adapt to semantic variations and robustly operate across different visual contexts and resolutions. The model's efficacy was evident in its ability to handle a wide range of object scales and orientations, making it a versatile tool for transportation infrastructure analysis. Moreover, its performance remained stable under varying lighting conditions and occlusions, which are common challenges in real-world scenarios. These attributes underscore the model's potential for practical applications in intelligent transportation systems, where accurate and reliable detection of infrastructure components is critical for operational efficiency and safety. This robust performance suggests that OWL-ViT could significantly enhance the capabilities of existing traffic management and autonomous vehicle systems by providing more accurate and timely information about the surrounding environment.

Future work should focus on enhancing detection accuracy for challenging categories, conducting broader testing in diverse transportation scenarios, and exploring integration with other sensory inputs to create a comprehensive detection system. Investigating real-time implementation will also be crucial to ensure the model's practicality in dynamic environments.

## References

1.    Zhai, F., Luan, J., Xu, Z., Chen, W.: DetReco: Object-Text Detection and Recognition Based on Deep Neural Network. Math Probl Eng. 2020, 1–15 (2020). https://doi.org/10.1155/2020/2365076.

2. Liu, E.: Simplifying Multimodal Composition: A Novel Zero-Shot Framework to Visual Question Answering and Image Captioning. (2023). https://doi.org/10.21203/rs.3.rs-3027308/v1.
3. Liu, W., Sun, W., Liu, Y.: DLOAM: Real-Time and Robust LiDAR SLAM System Based on CNN in Dynamic Urban Environments. Ieee Open Journal of Intelligent Transportation Systems. 1–1 (2024). https://doi.org/10.1109/ojits.2021.3109423.
4. Rozsa, Z., Szirányi, T.: Object Detection From a Few LIDAR Scanning Planes. Ieee Transactions on Intelligent Vehicles. 4, 548–560 (2019). https://doi.org/10.1109/tiv.2019.2938109.
5. Khan, S.: Transformers in Vision: A Survey. (2021). https://doi.org/10.48550/arxiv.2101.01169.
6. Khan, W.A., Liu, F., Vos, M.: Improving Object Detection in Real-World Traffic Scenes. 288–299 (2021). https://doi.org/10.1007/978-3-030-72073-5_22.
7. Osokin, A., Sumin, D., Lomakin, V.: OS2D: One-Stage One-Shot Object Detection by Matching Anchor Features. (2020). https://doi.org/10.48550/arxiv.2003.06800.
8. Michaelis, C., Bethge, M., Ecker, A.: A Broad Dataset Is All You Need for One-Shot Object Detection. (2020). https://doi.org/10.48550/arxiv.2011.04267.
9. Huang, Q., Zhang, H., Xue, M., Song, J., Song, M.: A Survey of Deep Learning for Low-Shot Object Detection. (2021). https://doi.org/10.48550/arxiv.2112.02814.
10. Biswas, S.K., Milanfar, P.: One shot detection with laplacian object and fast matrix cosine similarity. IEEE Trans Pattern Anal Mach Intell. 38, 546–562 (2015).
11. Wei, Y., Ma, Y.: Zero-Shot Object Detection With Multi-Label Context. (2022). https://doi.org/10.18293/seke2022-012.
12. Bansal, A., Sikka, K., Sharma, G., Chellappa, R., Divakaran, A.: Zero-Shot Object Detection. 397–414 (2018). https://doi.org/10.1007/978-3-030-01246-5_24.
13. Rahman, S., Khan, S.A., Barnes, N., Khan, F.S.: Any-Shot Object Detection. (2020). https://doi.org/10.48550/arxiv.2003.07003.
14. Cho, H.-C., Jhoo, W.Y., Kang, W., Roh, B.: Open-Vocabulary Object Detection Using Pseudo Caption Labels. (2023). https://doi.org/10.48550/arxiv.2303.13040.
15. Qian, H., Mei, A., Xu, H., Chen, J., Zhou, X., Liu, Y., Min, X., Yang, H., Li, X., Yang, J., Zhong, J., Jiang, X., Li, D.: Association of Serum Uric Acid and Blood Pressure: A Cross-Sectional Study in a Working Population in China. (2022). https://doi.org/10.21203/rs.3.rs-1308183/v1.
16. Dosovitskiy, A., Beyer, L., Kolesnikov, A., Weissenborn, D., Zhai, X., Unterthiner, T., Dehghani, M., Minderer, M., Heigold, G., Gelly, S.: An image is worth 16x16 words: Transformers for image recognition at scale. arXiv preprint arXiv:2010.11929. (2020).

17. Radford, A., Kim, J.W., Hallacy, C., Ramesh, A., Goh, G., Agarwal, S., Sastry, G., Askell, A., Mishkin, P., Clark, J.: Learning transferable visual models from natural language supervision. In: International conference on machine learning. pp. 8748–8763. PMLR (2021).
18. Carion, N., Massa, F., Synnaeve, G., Usunier, N., Kirillov, A., Zagoruyko, S.: End-to-end object detection with transformers. In: European conference on computer vision. pp. 213–229. Springer (2020).
19. Carion, N., Massa, F., Synnaeve, G., Usunier, N., Kirillov, A., Zagoruyko, S.: End-to-end object detection with transformers. In: European conference on computer vision. pp. 213–229. Springer (2020).